\documentclass[10pt,twocolumn,letterpaper]{article}

\usepackage[pagenumbers]{cvpr} 

\usepackage{graphicx}
\usepackage{amsmath}
\usepackage{amssymb}
\usepackage{booktabs}
\usepackage{algorithmic}
\usepackage{algorithm}
\usepackage{comment}
\usepackage[inline]{enumitem}
\usepackage{soul}
\usepackage{mathtools}
\usepackage{multirow}
\usepackage{tabularx}
\newcolumntype{Y}{>{\centering\arraybackslash}X}

\usepackage[pagebackref,breaklinks,colorlinks]{hyperref}

\usepackage[capitalize]{cleveref}
\crefname{section}{Sec.}{Secs.}
\Crefname{section}{Section}{Sections}
\Crefname{table}{Table}{Tables}
\crefname{table}{Tab.}{Tabs.}

\def\cvprPaperID{3} 
\def\confName{CVPR}
\def\confYear{2023}

\begin{document}

\title{Peer-to-Peer Federated Continual Learning for Naturalistic Driving Action Recognition}

\author{Liangqi Yuan, Yunsheng Ma, Lu Su, Ziran Wang\\
Purdue University, West Lafayette, USA\\
{\tt\small \{liangqiy, yunsheng, lusu, ziran\}@purdue.edu}\\
}

\maketitle

\begin{abstract}
Naturalistic driving action recognition (NDAR) has proven to be an effective method for detecting driver distraction and reducing the risk of traffic accidents. However, the intrusive design of in-cabin cameras raises concerns about driver privacy. To address this issue, we propose a novel peer-to-peer (P2P) federated learning (FL) framework with continual learning, namely FedPC, which ensures privacy and enhances learning efficiency while reducing communication, computational, and storage overheads. Our framework focuses on addressing the clients' objectives within a serverless FL framework, with the goal of delivering personalized and accurate NDAR models. We demonstrate and evaluate the performance of FedPC on two real-world NDAR datasets, including the State Farm Distracted Driver Detection and Track 3 NDAR dataset in the 2023 AICity Challenge. The results of our experiments highlight the strong competitiveness of FedPC compared to the conventional client-to-server (C2S) FLs in terms of performance, knowledge dissemination rate, and compatibility with new clients. 
\end{abstract}

\section{Introduction}
\label{Sec:Introduction}

\begin{figure}[t]
\centering
\centerline{\includegraphics[width=0.6\linewidth]{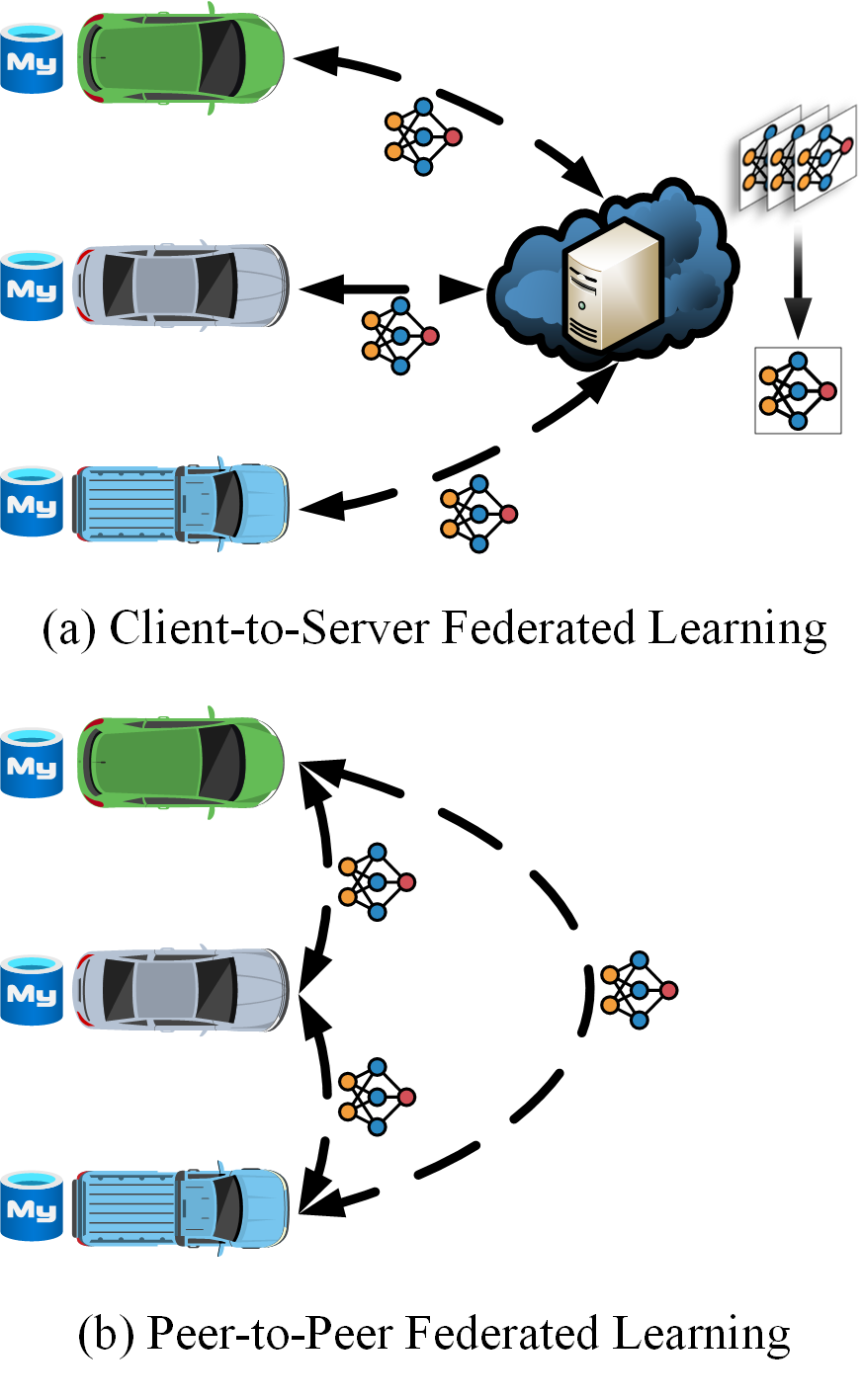}}
\caption{Structures of two FL frameworks, (a) C2S FL and (b) P2P FL.}
\label{Fig. C2S_vs_P2P}
\end{figure}

\begin{table*}[t] 
  \small
  \centering
  \caption{FedPC vs C2S FL.}
  \begin{tabularx}{\linewidth}{l|Y|Y}
    \toprule
    System & C2S FL (FedAvg) & FedPC (proposed) \\
    \midrule
    Objective & Clients: a personalized model for each client. \par Server: a single generalized model & Clients: a personalized model for each client. \par Server: N/A \\
    \midrule
    Knowledge Dissemination & Server aggregation and transmission & Continual learning from another client model \\
    \midrule
    Communication Complexity & Client: send $1$ model per iteration round \par Server: send $|C|$ models per iteration round & Client: send $1$ model per iteration round \par Server: N/A \\
    \midrule
    Dissemination Rate & Slow, it needs to wait for the server to receive, aggregate, and transmit the models & Quick, it only requires clients to transmit the model to each other \\
    \midrule
    Generalizability & Stronger in IID datasets & Partial generalization with non-IID datasets \\
    \midrule
    Compatibility with New Clients & Poor, can be enhanced by personalization & Poor, personalization process may be faster \\
    \midrule
    Hardware Overhead & High, it requires server communication, computing and storage resources & Low \\
    \midrule
    Hidden Concern & Privacy breach, security, trust, SPoF, and aggregation fairness on the server & Lack of incentives, security, and deadlocks on the clients \\
    \bottomrule
  \end{tabularx}
  \label{Tab:FedPC vs C2S FL}
\end{table*}

Naturalistic driving action recognition (NDAR) is a crucial technology for ensuring road safety and reducing the risk of traffic accidents. By monitoring and characterizing driver behavior through biological information, NDAR aims to detect driver distractions, classify driver activities, and predict vehicle trajectories. Driver distraction remains a major cause of automobile accidents~\cite{zhang2019driving}. While driver monitoring applications are moving towards multimodal options~\cite{koesdwiady2016recent}, computer vision (CV) remains the most popular methodology for driver monitoring applications due to its maturity and high-performance system designs. However, the intrusive nature of CV and its potential for privacy breaches are major drawbacks~\cite{tom2017designing}. Furthermore, CV techniques often suffer from inefficiency in the learning process due to the large size of data involved~\cite{nishio2021wireless}. 

To address these challenges, federated learning (FL) offers a promising solution by sharing model weights instead of user data, ensuring the privacy of user information. The FL approach also reduces the need for communication resources as the size of the model parameters is usually much smaller than the user data. FL has been demonstrated to improve not only the learning efficiency of the system but also the model's generalization ability. It is widely used in various applications and is considered a paradigm for worldwide cooperation~\cite{dayan2021federated,pati2022federated}.

Centralized FL, also known as server-to-client (C2S) FL, started with FedAvg~\cite{mcmahan2017communication} and has evolved into a diverse range of approaches aimed at solving various challenges, including heterogeneity~\cite{li2020federated}, communication overhead~\cite{konevcny2016federated}, differential privacy~\cite{wei2020federated}, personalization~\cite{tan2022towards}, fairness~\cite{li2019fair}. In C2S FL, there exist two participants, namely client and server, with different objectives~\cite{shen2020federated}. The client seeks to obtain a personalized local model, while the server seeks a generalized global model. It is a challenge to balance the purpose between the client and the server. Decentralized FL, also known as peer-to-peer (P2P) FL, is a novel approach for knowledge dissemination where clients communicate directly with each other, bypassing the necessity for centralized server distribution, aggregation, and management~\cite{kairouz2021advances,rieke2020future}. \cref{Fig. C2S_vs_P2P} shows the illustration of C2S FL and P2P FL frameworks. 

P2P FL effectively emphasizes the clients' objectives due to the absence of the central server objective and competing relationships~\cite{belal2022pepper}. This results in a more resource-efficient and simpler communication process, reducing the single-point of failure (SPoF) risk associated with the centralized server. P2P FL utilizes network topology for client interconnection, making it a highly customizable solution. The design of the communication protocol, iteration order, and temporal variability are optional features that can be tailored to the specific requirements of the application scenario, leading to improved accuracy, robustness, and convergence~\cite{chellapandi2023convergence}. P2P FL offers a flexible framework for decentralized knowledge dissemination, allowing for custom topologies, communications, and iteration strategies to be devised based on the a priori information of the application scenario.

The way to propagate knowledge among clients and iteratively update models is an open problem in P2P FL. Without the coordination and aggregation capabilities of a central server, clients broadcast local models and store and aggregate models from other clients, resulting in more communication, computational, and storage overheads. Incremental learning, also known as continual learning, has gained attention through the paradigm of sustainable learning to accumulate prior knowledge and overcome catastrophic forgetting~\cite{de2021continual}. Continual learning has gained prominence in P2P FL because it propagates knowledge directly in the client without relying on average computation for knowledge aggregation~\cite{chang2018distributed,sheller2019multi,sheller2020federated,huang2022continual}. Furthermore, continual learning aligns with the real-world nature of infinite data streams generated by the client, thereby further avoiding the impact of concept drift on the model.

In this paper, we introduce a novel P2P FL framework, FedPC, for the Internet of Vehicles (IoV), aiming to address the challenges in multiple NDAR tasks. The proposed FedPC employs the continual learning paradigm in combination with a gossip protocol to propagate knowledge between clients and perform iterative model updates. \cref{Tab:FedPC vs C2S FL} presents a comparison of the proposed FedPC and C2S FL, including objectives, knowledge dissemination methods, communication complexity, etc. The objective of FedPC is to provide a solution that strikes a balance between privacy protection, learning efficiency, and generalizability to achieve accurate and effective NDAR. Experiments on two datasets, State Farm Distracted Driver Detection (StateFarm)~\cite{farm_2016} and Track 3 NDAR dataset in 2023 AICity Challenge (AICity)~\cite{Naphade23AIC23,Rahman22SynDD2}, demonstrate the results of FedPC in terms of performance, convergence, and compatibility with new clients. The contributions of this paper are:
\begin{itemize}
\item We propose a P2P FL system that addresses clients' objectives and incorporates a continual learning paradigm. Our system aims to reduce communication, computational, and storage overheads, improve the dissemination rate, and resolve issues associated with the server once and for all. To the best of our knowledge, this is one of the first papers that introduces a P2P FL system, combined with a continual learning framework, into the IoV.
\item We emphasize the characteristics of vehicle connectivity in real-world scenarios, including high dynamism, randomness, data heterogeneity, as well as communication, computational, and storage resource constraints. Through extensive simulation of application scenario experiments, we showcase the potential feasibility of deploying the proposed FedPC in real-world IoV environments.
\item We compare FedPC with two conventional C2S FL and ring P2P FL approaches using three evaluation metrics on two real-world NDAR datasets. The results reveal the proposed FedPC's strong emphasis on clients' objectives, exceptional performance, efficient knowledge dissemination rate, comparable generalizability, and rapid compatibility with new clients.
\end{itemize}

The presentation of this paper is as follows. \cref{Sec:Related Works} reviews related works of P2P FL and FL for connected vehicles. The problem formulation and proposed solution are described in \cref{Sec:Methodology}. The datasets, implementations, and results are demonstrated in \cref{Sec:Experiment}. \cref{Sec:Discussion} discusses potential deployments of the proposed FedPC in real-world application scenarios, followed by \cref{Sec:Conclusion} summarizing the paper and expounding on future work.

\section{Related Works}
\label{Sec:Related Works}

\subsection{Peer-to-Peer Federated Learning}
\noindent \textbf{Network Topology.} Communication network topologies are fundamental to P2P FLs, as they determine the communication protocol and knowledge dissemination process of the network. Some common network topologies include line, ring, mesh, and star, each designed to address communication, computational, convergence, and other challenges. Hybrid topologies offer highly customizable structures that can adapt to various application fields. Shi \textit{et al.}~\cite{shi2021over} proposed a hybrid P2P FL and demonstrated the performance in convergence. Wang \textit{et al.}~\cite{wang2022matcha} proposed a dynamic hybrid P2P FL, Matcha, which aims to balance convergence speed and communication complexity. Matcha improves the convergence speed by giving higher communication frequency to key clients while reducing communication delay by decreasing the communication frequency of other clients.

\noindent \textbf{Paradigm: Aggregation and Continual.} Compared to C2S FL, P2P FL has no centralized server for coordination and distribution, making the integration of knowledge from all clients an open problem. Aggregating all model parameters is challenging, as determining where the aggregation occurs raises security, privacy, and fairness concerns. Chen \textit{et al.}~\cite{chen2022decentralized} proposed a decentralized FL framework, in which each client broadcasts its local model to all other clients, and each client performs local aggregation operations using the models received from others. Roy \textit{et al.}~\cite{chen2022decentralized} proposed a FL framework, BrainTorrent, where the current client requests models from other clients and then performs the aggregation operation locally upon receiving them. However, these aggregation-based approaches invariably require higher model transmission frequency and model storage requirements per client. Some P2P FL frameworks employ the continual paradigm to retain previous clients' knowledge while incrementally and continuously learning using the same model. Feasibility studies and preliminary experiments on line and ring topologies have been conducted \cite{chang2018distributed,sheller2019multi,sheller2020federated,huang2022continual}. Nonetheless, for sequential topologies like line and ring, the convergence of the system strongly heavily depends on the clients' iteration order. Also, the local convergence of the subsequent client strongly relies on the performance of the prior model. The speed of iteration, SPoF, privacy, incentives, and other factors are challenges.

\subsection{Federated Learning for Connected Vehicles}
\noindent \textbf{FL for NDAR.} FL has been one of the solutions for NDAR tasks due to its protection of driver privacy, knowledge dissemination strategy, and adaptability to application scenarios~\cite{chellapandi2023survey}. Various FL variants have been proposed for adaptation to NDAR. Doshi \textit{et al.}~\cite{doshi2022federated} proposed a FedGKT-based~\cite{he2020group} framework to transfer small model knowledge from in-vehicle edge devices to a large server and integrate it into a large model through knowledge distillation. Yuan \textit{et al.}~\cite{yuan2023federated} proposed a FL framework, known as FedTOP, which combines transfer, ordered, and personalized modules to address communication overhead, security, and heterogeneity issues. Within the FedTOP framework, two datasets with varying degrees of data heterogeneity are compared.

\noindent \textbf{Vehicle Connection in P2P FL.}
P2P FL frameworks are becoming increasingly important for enabling connected vehicles to learn from each other without compromising data privacy. These frameworks serve a variety of purposes and have been proposed in various forms. For example, Nguyen \textit{et al.}~\cite{nguyen2022deep} proposed a ring P2P FL framework for autonomous driving vehicles. Yu \textit{et al.}~\cite{yu2020proactive} proposed a star topology framework, in which a vehicle acts as a client while also assuming the responsibilities of a server. Lu \textit{et al.}~\cite{lu2020federated} relied on roadside units (RSUs) as the corresponding relay station to broadcast the vehicle request information to neighboring vehicles. The requesting vehicle receives the models directly from the responding neighboring vehicles and aggregates them locally.

\begin{figure*}[t]
\centering
\centerline{\includegraphics[width=1\linewidth]{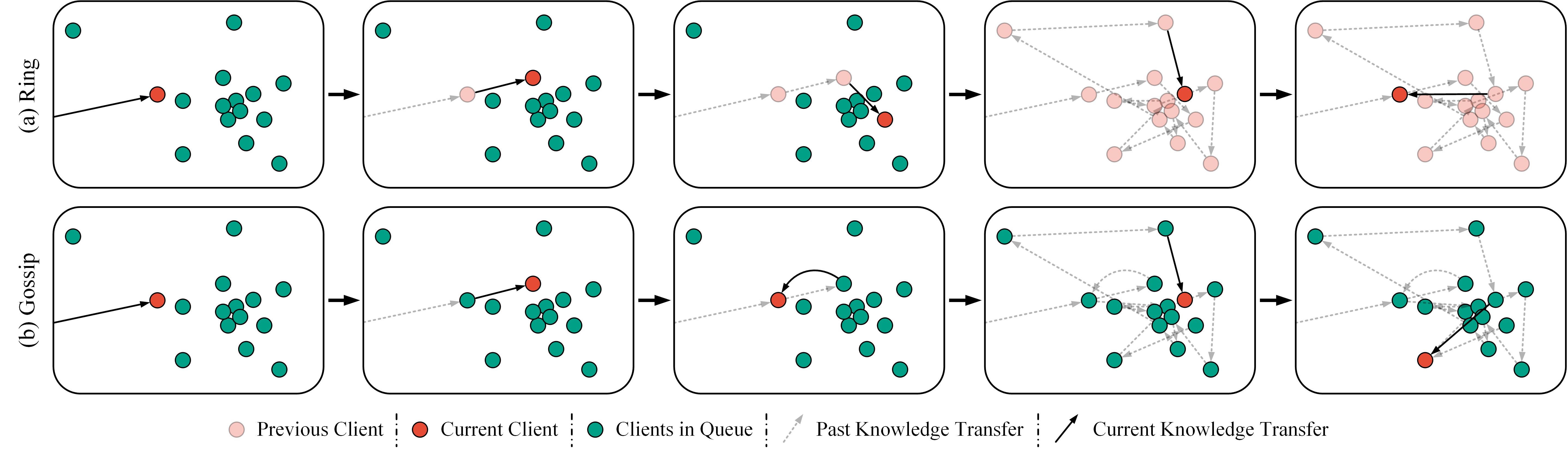}}
\caption{Model iteration process in P2P FL networks for (a) ring and (b) gossip communication protocols. Due to statistical heterogeneity and system heterogeneity, client models in the parameter space are non-IID.}
\label{Fig. Parameter Space}
\end{figure*}

\section{Methodology}
\label{Sec:Methodology}

\subsection{Problem Formulation}
In FL, data are isolated for each client, and the only parameters propagated between clients are the model parameters. There exists a set with $C$ clients, and for each client $c$ there exists an nested local model $\omega_c$ such that
\begin{equation}
    Y_c = \omega_c(X_c),
\label{Eq. nested model}
\end{equation}
where $X_c$ and $Y_c$ are the isolated local data set and the label set, respectively. For FedPC, there is no global objective, but only local objective for each client
\begin{equation}
    \min_{\omega} \mathcal{L}(\omega; X_c, Y_c),
\label{Eq. objective}
\end{equation}
where $\mathcal{L}$ is a loss function. 

We assume that the optimal client models $\omega^\ast$ obey a multivariate distribution, which can be considered as a superposition of multiple multivariate distributions due to the non-IID of the data and system heterogeneity. For continual learning, the initial model parameters for the current client will be the previous client model. Therefore, the optimization process of the current client model can be expressed as
\begin{equation}
    \omega_c = \arg\min_{\omega} \mathcal{L}(\omega;X_c, Y_c, \omega_{c-1}).
\label{Eq. optimization}
\end{equation}

\cref{Eq. optimization} can be considered as a Bayesian model, i.e., the probability distribution of the posterior model is strongly based on the probability distribution of the prior model. In addition, the posterior model is determined by order of the models, the distance between the prior and the posterior models, the merit of the training, and so on.

\subsection{Communication Protocol}
\label{Sec:Communication Protocol}
The communication protocol used in FedPC determines the order of model iterations, which in turn affects the knowledge dissemination process, the distance between prior and posterior models, and the system robustness. In this paper, we propose using the gossip protocol for communication between clients instead of the ring P2P FL for the following reasons. Firstly, the introduction of randomness adds robustness to the system~\cite{alvisi2007robust}. Secondly, ring P2P FLs need to assume a certain priori knowledge, such as that each client knows the information about previous and subsequent clients. Thirdly, vehicle-to-vehicle (V2V) communication is defined by vehicular ad hoc networks (VANETs)~\cite{khelifi2019named}, e.g., dedicated short-range communications (DSRC) among on-board units (OBU). VANET is highly dynamic and limited by transmission range, so the stationary between the ring vehicle connections cannot be guaranteed. Fourthly, there are SPoF concerns for unidirectional connected ring topology networks. 

\cref{Fig. Parameter Space} shows the ring P2P FL and the proposed FedPC with gossip protocol. Each client in a ring P2P FL knows the information of the previous and subsequent clients, and each client is propagated only once during each iteration. In contrast, for the gossip protocol, each client is unsure of the information of previous and subsequent clients, and each client may not be propagated or may be propagated multiple times during each iteration. 

\subsection{Training Strategy}
\label{Sec:Training Strategy}
The order of client iterations will seriously affect the performance of the client model and the convergence rate of the whole system. Each client model will strongly depend on the performance of the previous client model. We consider several training strategies to accelerate convergence, enhance robustness, and improve accuracy.

\noindent \textbf{Proximal Term Loss.} Although the clients' objectives are to obtain a personalized model, we do not want the local model to be overfitted on the client's data for the training iteration of the system. Therefore, a loss function is needed to penalize the distance of the current model from the prior model. $\mathcal{L}$ consists of two parts, including a negative log-likelihood (NLL) loss $\mathcal{L}_{nll}$ and a proximal term $\mathcal{L}_p$~\cite{li2020federated}. $\mathcal{L}_{nll}$ is used to calculate the difference between the true and predicted values, while $\mathcal{L}_p$ is used to penalize the distance between the current client model and the previous client model. The overall loss function can be expressed as
\begin{equation}
\begin{split}
    \mathcal{L} &= \mathcal{L}_{nll} + \mathcal{L}_p, \\
    \mathcal{L} &= \mathcal{L}_{nll} + \frac{\mu}{2}\|\omega_{c}-\omega_{c-1}\|^2,
\end{split}
\label{Eq. loss function}
\end{equation}
where $\mu$ is the proximal term penalty factor, which we set to $1$ to be the same as in other studies.

\noindent \textbf{Transfer Learning.} Due to the communication overhead caused by the transmission of model parameters in FL, transfer learning reduces the communication overhead while enhancing model generalization and avoiding overfitting by freezing the low-level parameters of the model. Due to the lack of a generalized NDAR model, we use ResNet34~\cite{he2016deep} pre-trained on ImageNet as the base model and freeze the parameters of the first three ResNet blocks. Although there are significant differences between ImageNet and NDAR tasks, the low-level layers of the CNN can be considered as a feature extractor without losing generality. The original model size is 83 MB, but by implementing the transfer learning paradigm, the model size is reduced to 52 MB, resulting in a 37\% decrease in communication overhead. 

\begin{algorithm}[t]
\small
   \caption{\small{Proposed peer-to-peer federated learning frame-
work with continual learning (FedPC)}}
   \label{Alg FedPC}
\begin{algorithmic}[1]
   \renewcommand{\algorithmicrequire}{\textbf{Input:}}
   \renewcommand{\algorithmicensure}{\textbf{Output:}}
   \REQUIRE Iteration rounds ($T$), client set ($C$), data set ($X_c$) and label set ($Y_c$) for each client ($c \in C$), local training epoch ($E$), initial model ($\omega_0$), loss function ($\mathcal{L}$), learning rate ($\eta_t$)
   \ENSURE Trained local models for each client ($\{\omega_{c} | c \in C\}$)
   \FOR{$t=1$ {\bfseries to} $T-1$}
        \FOR{$c \in C$ {\bfseries in gossip}}
            \STATE Receive the model parameters sent by the previous client $\omega_c \gets \omega_{c-1}$.
            \FOR{$e=1$ {\bfseries to} $E-1$}
                \STATE Backpropagate the loss function and update the local model $\omega_c^{e+1} \gets \arg\min_{\omega_c^{e}} \mathcal{L}(\omega_c^{e})$.
            \ENDFOR
            \STATE Update the local model $\omega_c \gets \omega_c^{E}$.
            \STATE Client $c$ gossip $\omega_c$ to the next client $c+1$.
        \ENDFOR
   \ENDFOR
\end{algorithmic}
\end{algorithm}

\noindent \textbf{Decreasing Learning Rate.} The learning rate is also one of the essential hyperparameters in continual learning. Too large a learning rate will cause catastrophic forgetting of the system, while too small a learning rate will cause the model to fail to learn the client's knowledge. Therefore, we consider a learning rate $\eta_t$ decreasing strategy based on the iteration. A higher learning rate is assigned at the initial stage of the iteration to enable the edge model to slim down quickly and a lower learning rate in subsequent iterations to avoid catastrophic forgetting. 

We summarize the steps of FedPC in \cref{Alg FedPC}.

\begin{figure*}[t]
\small
\centering
\renewcommand{\tabcolsep}{0pt}
\begin{tabular}{cccccc}
\rotatebox[origin=c]{90}{(a) StateFarm} \hspace{6pt} &
\raisebox{-0.5\height}{\includegraphics[width=0.19\linewidth]{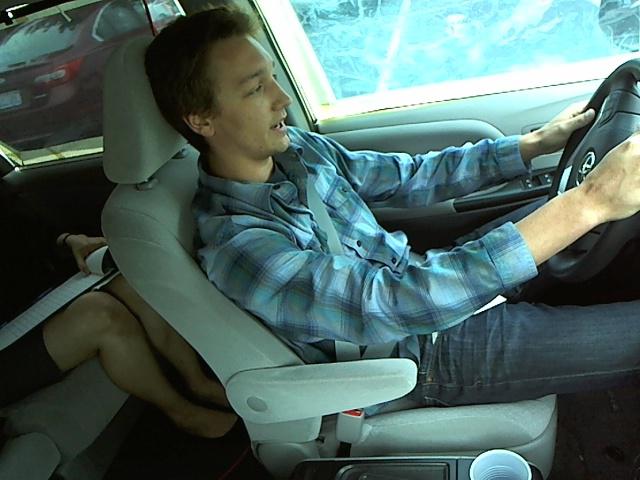}%
\label{Fig. SF1}} &
\raisebox{-0.5\height}{\includegraphics[width=0.19\linewidth]{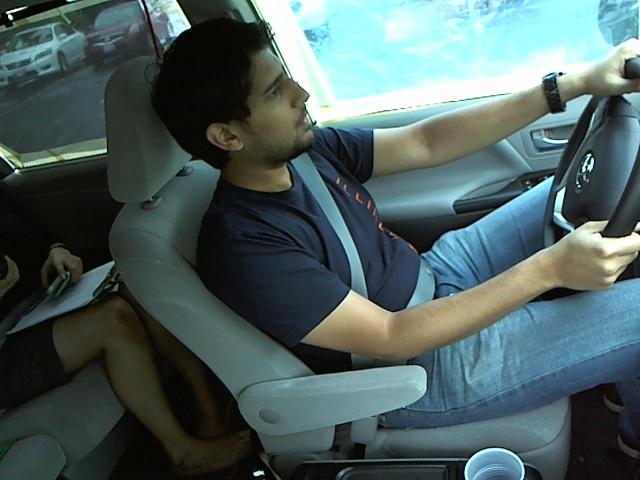}%
\label{Fig. SF2}} &
\raisebox{-0.5\height}{\includegraphics[width=0.19\linewidth]{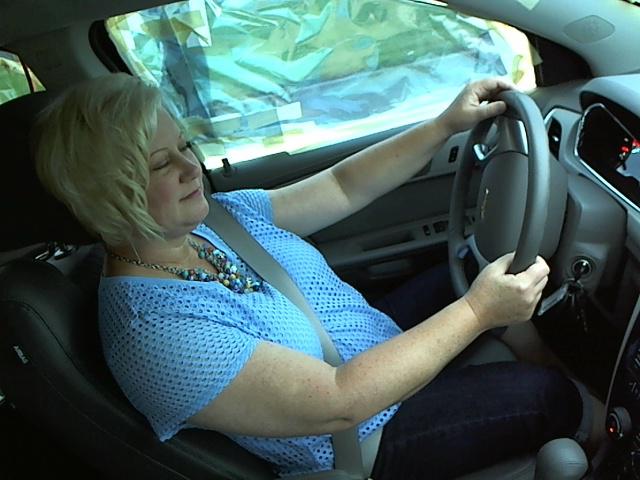}%
\label{Fig. SF3}} &
\raisebox{-0.5\height}{\includegraphics[width=0.19\linewidth]{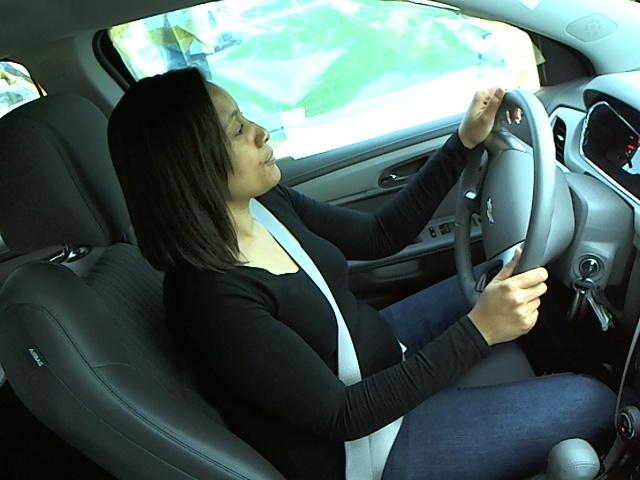}%
\label{Fig. SF4}} &
\raisebox{-0.5\height}{\includegraphics[width=0.19\linewidth]{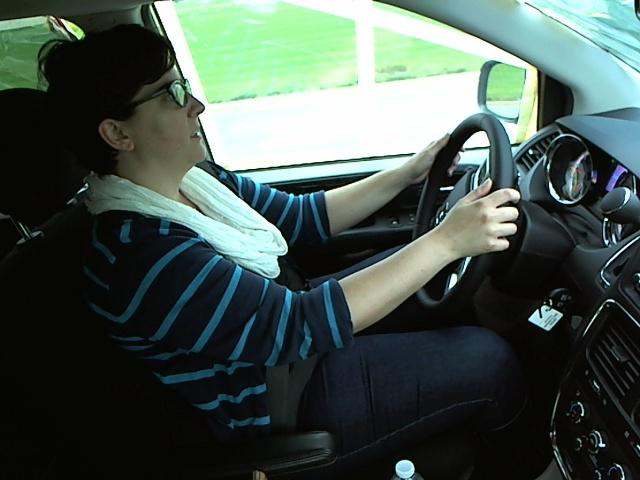}%
\label{Fig. SF5}} \\
\rotatebox[origin=c]{90}{(b) AICity} \hspace{6pt} &
\raisebox{-0.5\height}{\includegraphics[width=0.19\linewidth]{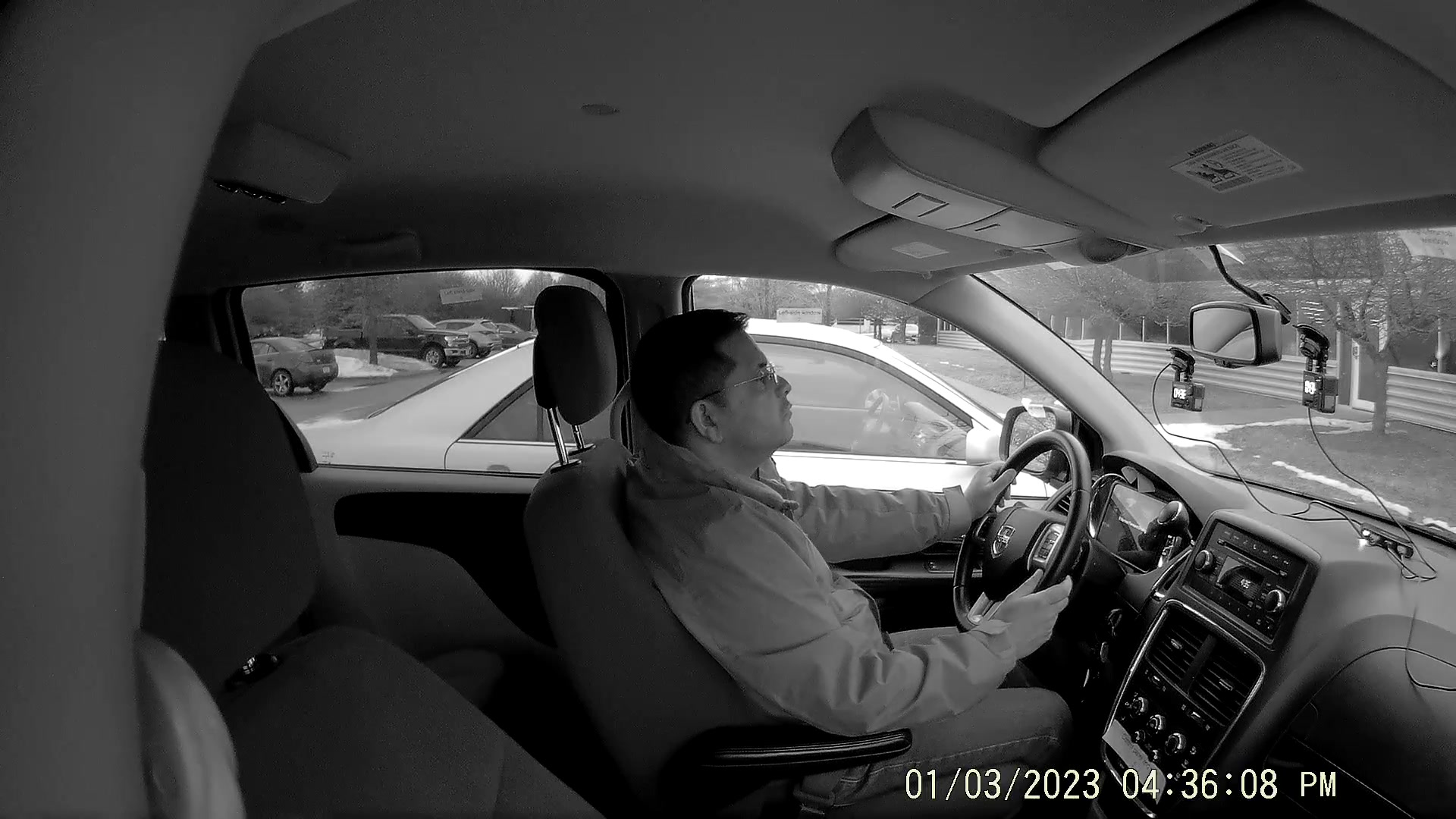}%
\label{Fig. AICity1}} &
\raisebox{-0.5\height}{\includegraphics[width=0.19\linewidth]{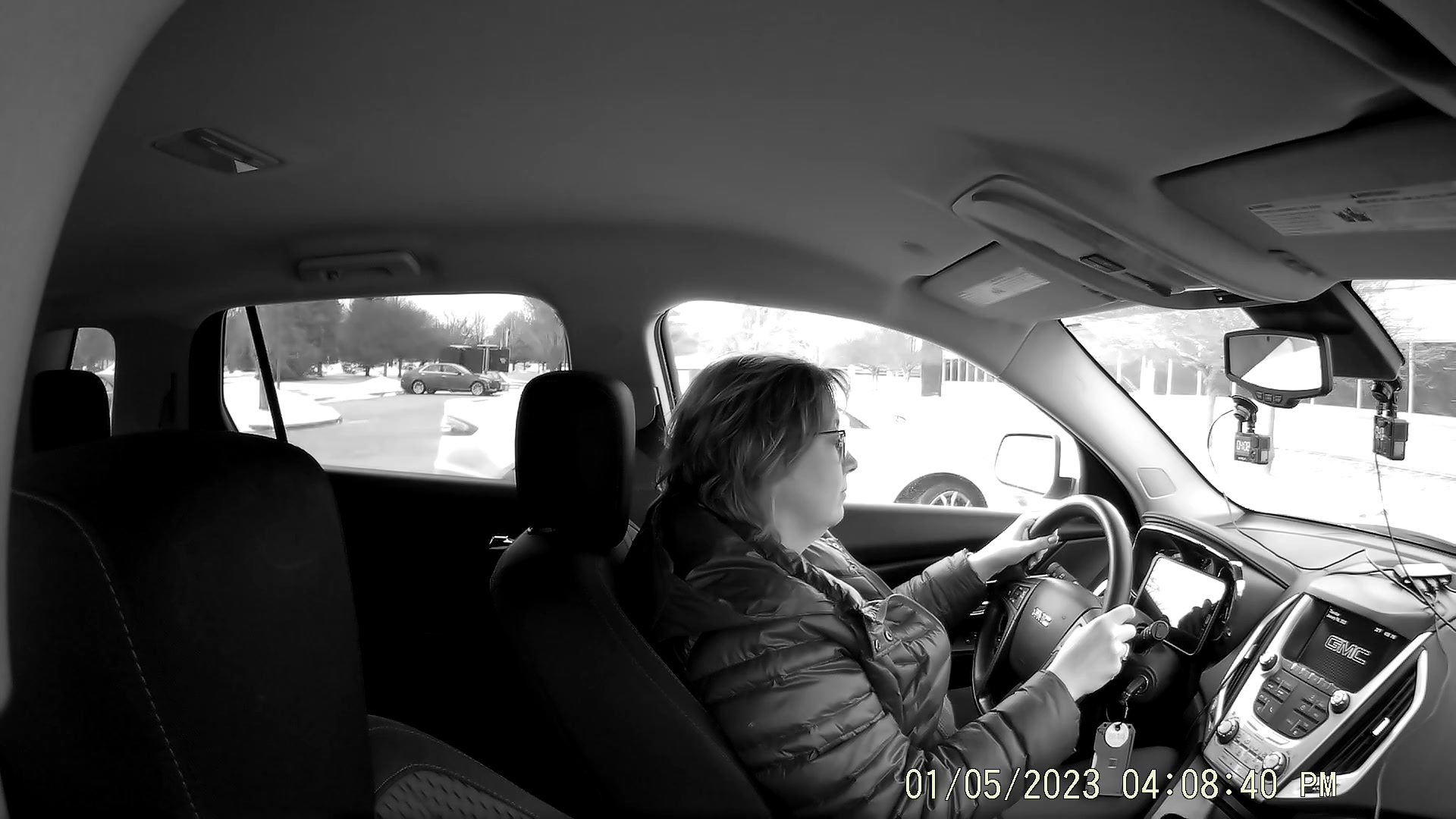}%
\label{Fig. AICity2}} &
\raisebox{-0.5\height}{\includegraphics[width=0.19\linewidth]{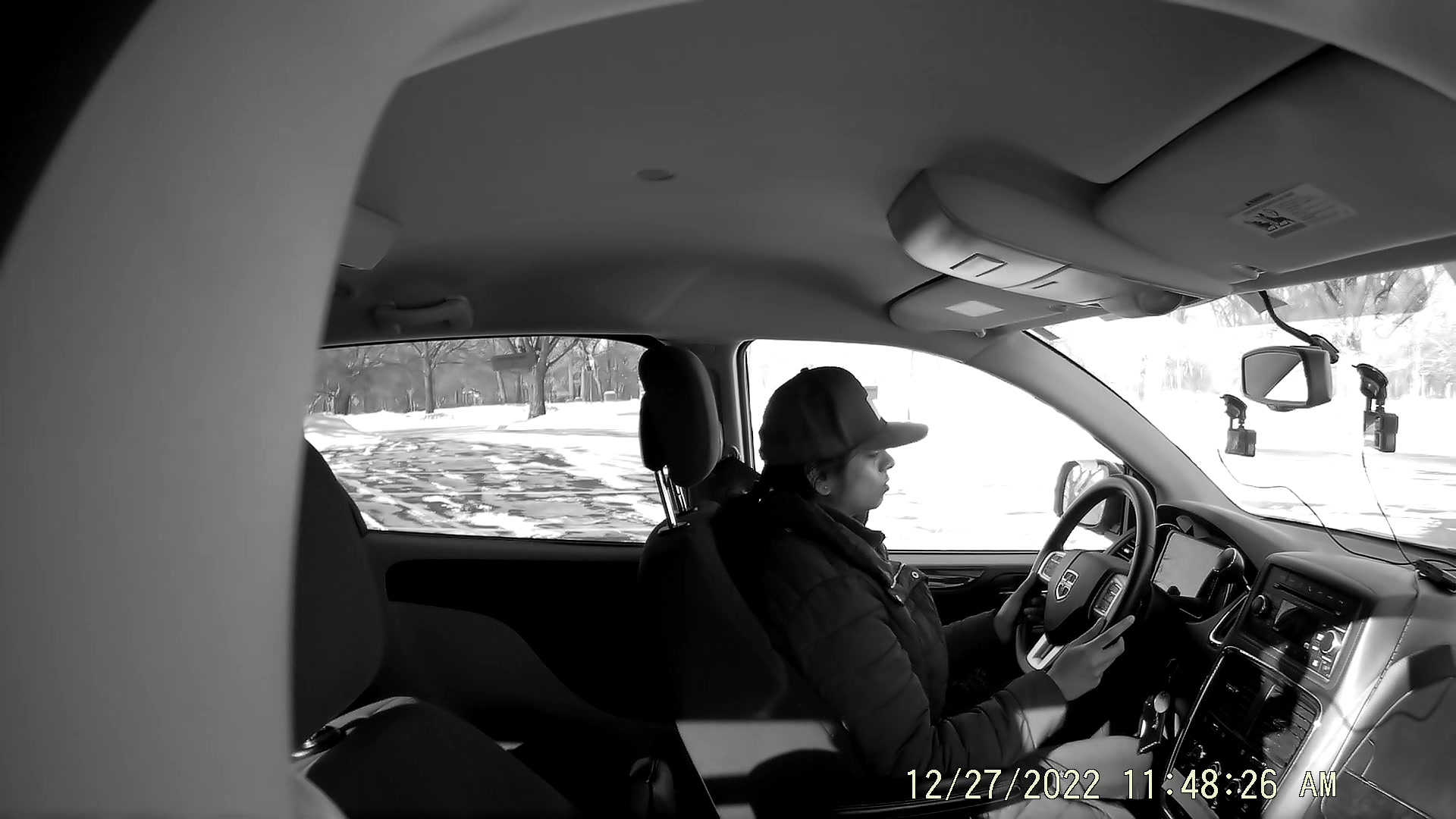}%
\label{Fig. AICity3}} &
\raisebox{-0.5\height}{\includegraphics[width=0.19\linewidth]{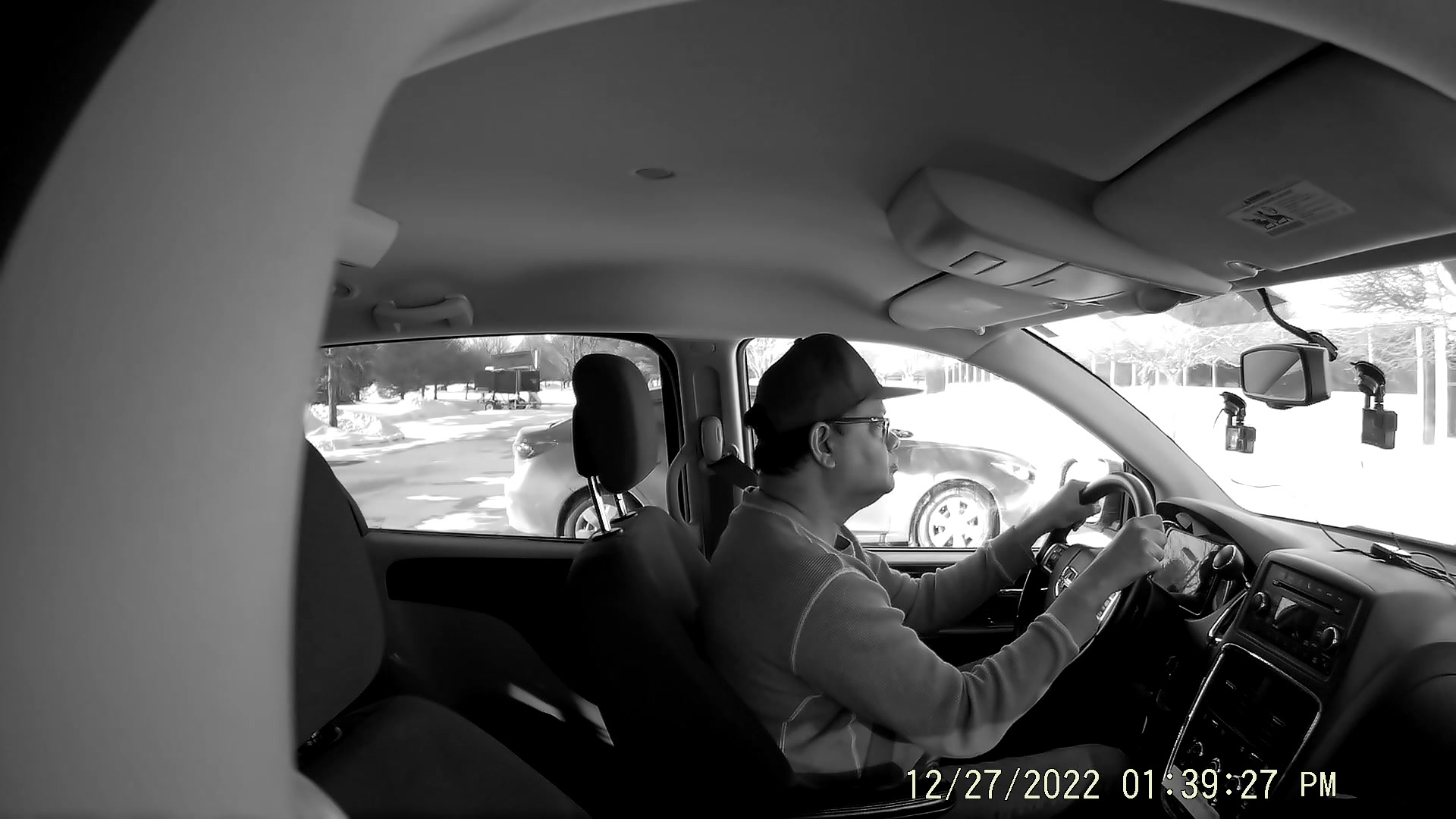}%
\label{Fig. AICity4}} &
\raisebox{-0.5\height}{\includegraphics[width=0.19\linewidth]{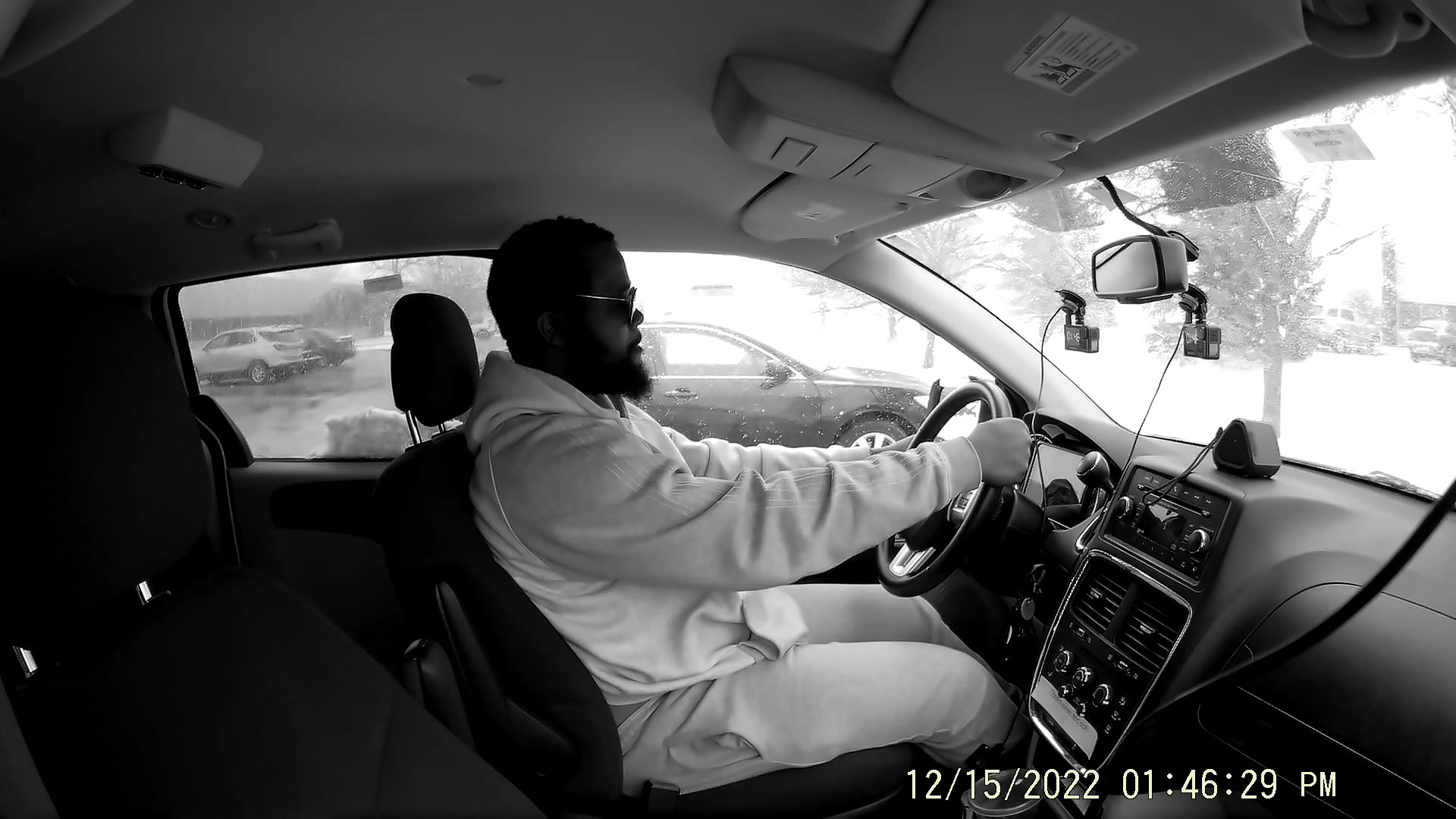}%
\label{Fig. AICity5}}
\end{tabular}
\caption{Two NDAR datasets are used in the experiments, including (a) StateFarm \cite{farm_2016} and (b) AICity \cite{Naphade23AIC23,Rahman22SynDD2} datasets.}
\label{Fig. Dataset} 
\end{figure*}

\section{Experiment}
\label{Sec:Experiment}

\subsection{Experimental Setup}

\noindent \textbf{StateFarm Dataset~\cite{farm_2016}.} As shown in \cref{Fig. Dataset} (a), the StateFarm dataset is a popular dataset in NDAR tasks due to its clean data, realistic on-road driving simulations, and diverse driving systems. In order to simulate a realistic driving environment, data collection is performed on vehicles being dragged by a truck traveling on the streets. It includes more than 100k image samples from 26 drivers in 10 categories. In particular, it includes multiple vehicles makes, interiors, and camera angles to emphasize the heterogeneity in real driving.

\noindent \textbf{AICity Dataset \cite{Naphade23AIC23,Rahman22SynDD2}.} As shown in \cref{Fig. Dataset} (b), the AICity dataset is a brand new, dense, and high-resolution dataset. It includes 34 hours of NDAR video footage from 35 drivers in 16 categories of distracted activity. It not only has three camera angles, including dashboard, rearview, and right side window, but also has different driver appearance blocks, including none, sunglasses, and hat. Data collection is also conducted on different vehicles with some minor camera adjustments. Unlike StateFarm, the AICity dataset does not collect data in a real driving environment but in a stopped state on different streets. In our experiment, each client's dataset consists of video frames captured from the right-side window camera angle of each driver.

\begin{figure}[t]
\centering
\subfloat[StateFarm]{\includegraphics[width=0.5\linewidth]{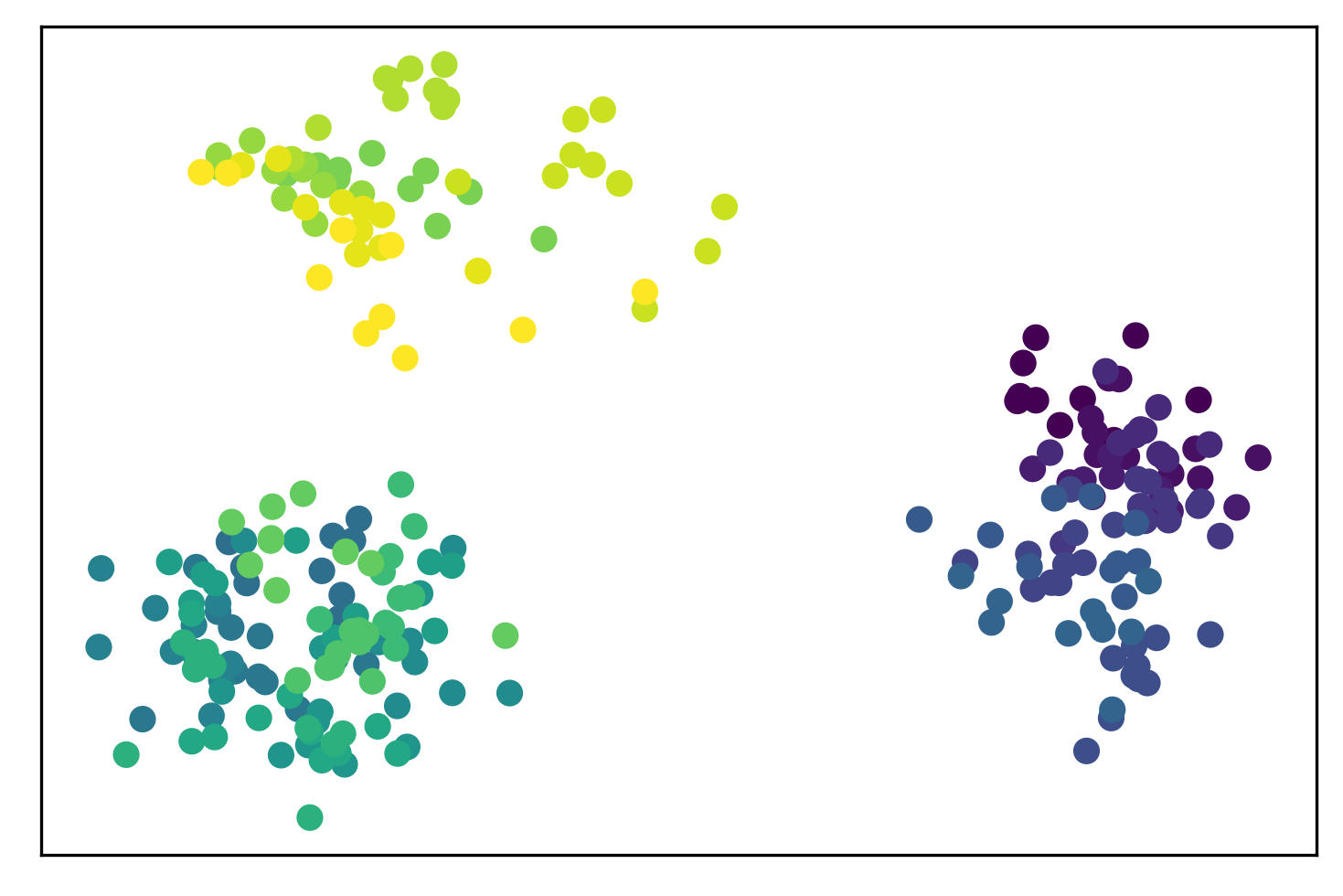}%
\label{Fig. StateFarm PCA}}
\hfill
\subfloat[AICity]{\includegraphics[width=0.5\linewidth]{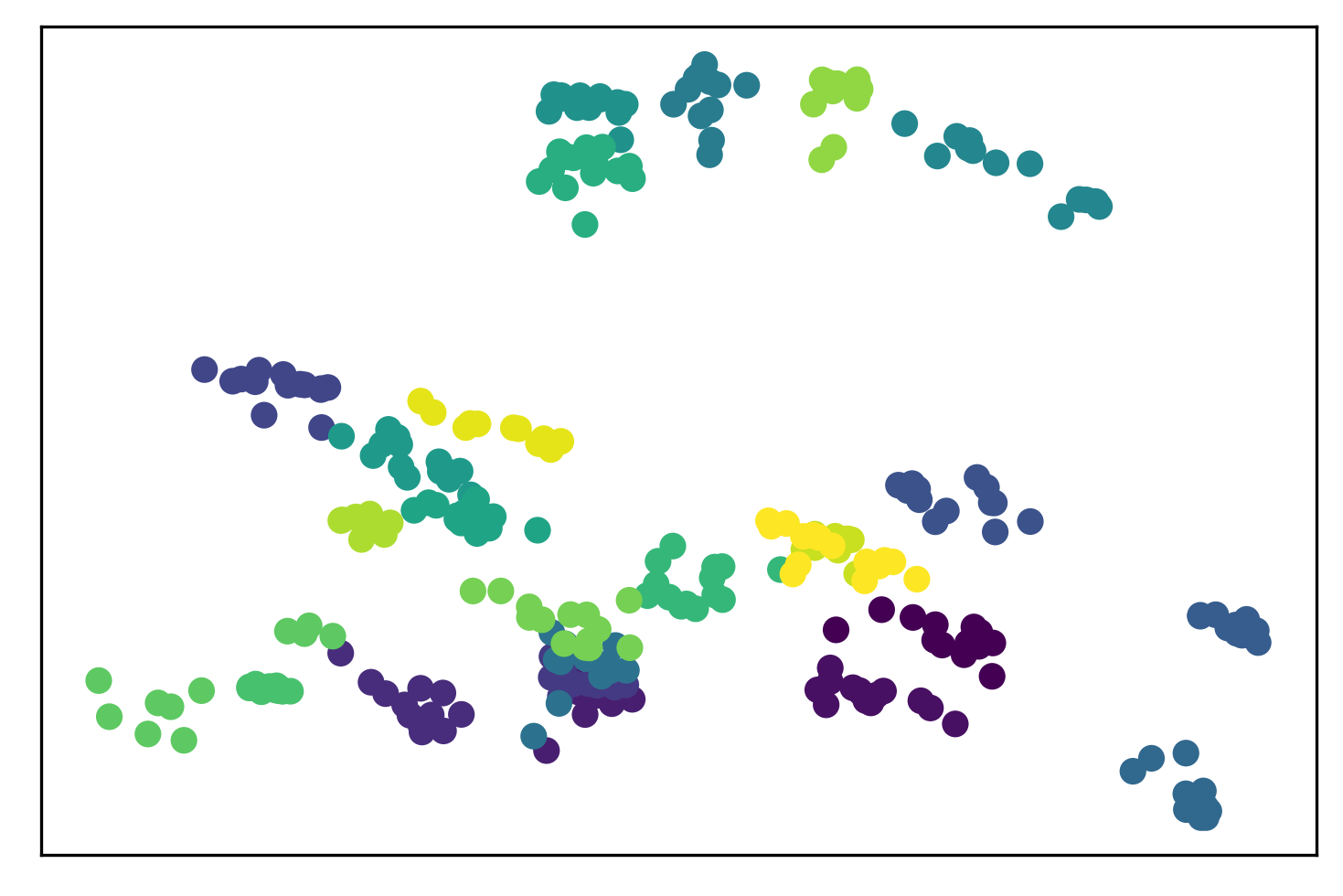}%
\label{Fig. AICity PCA}}
\caption{Data distributions of (a) StateFarm and (b) AICity. The colors denote different clients, and the scatter points represent different activities. The data distributions are visualized by averaging the driver activity images and reducing the dimensions through principal component analysis (PCA).}
\label{Fig. Dataset PCA} 
\end{figure}

\noindent \textbf{Understanding of Data Distribution.} NDAR application emphasizes the real and natural data distributions, as opposed to artificially creating non-IID settings as in~\cite{doshi2022federated}. Each client's data distribution exhibits both statistical and system heterogeneity. Statistical heterogeneity depends on factors such as the driver's physical characteristics, behaviors, and postures, while system heterogeneity depends on characteristics such as the vehicle model, interior, camera angle, and exterior environment. \cref{Fig. Dataset PCA} illustrates the data distribution of the StateFarm and AICity datasets. It can be seen that the StateFarm clients' data distribution is divided into three distinct clusters, which is consistent with our expectation since the dataset was collected from three different vehicles. Similarly, the AICity dataset is divided into two clusters due to the use of two different vehicles. In addition to the system heterogeneity introduced by the vehicles, the statistical heterogeneity of the drivers presents a challenge. Within the same cluster, the StateFarm client's data distribution is more overlapping, while the AICity client's data distribution is more dispersed. The dispersion and non-IID characteristics of the data distribution also lead to non-IID models in FL, which in turn creates compatibility issues between clients and a single FL model.

\noindent \textbf{Implementation.} In addition to the training strategy described in \cref{Sec:Training Strategy}, we set the number of iteration rounds $T = 5$, the local epoch $E = 5$, the batch size $128$, and the optimizer as Adam. The initial learning rate is $10^{-4}$, and the decay of each iteration round is 0.5 times. In addition to the two loss functions in \cref{Eq. loss function}, we set the weight decay to $10^{-5}$. We resize the image size of both datasets to 224 by 224 to avoid overfitting and to reduce storage and computational overhead.

We employ a double-splitting approach to divide the two aforementioned datasets into test sets. Firstly, we partition the clients into training and test clients using an 0.8 and 0.2 ratio. Secondly, we randomly split the training and test datasets for each client with the 0.8 and 0.2 ratios. Note that during the iterative training process, the test clients are not involved. However, it is possible to execute one or more gradient descents outside of the iterative process to personalize the model for the test clients. The experiments are conducted on a NVIDIA A100 GPU using the PyTorch framework.

\noindent \textbf{Baseline.} We consider the proposed FedPC for comparison on four baselines, including independent learning, FedAvg, FedProx, and ring P2P FL. Note that the line topology can be considered as the ring P2P FL with a single iteration, i.e., $T=1$. Owing to the potential confusion between FedPC and independent learning, resulting from the clients' objectives and absence of aggregation operation, we also incorporate independent learning as one of the baseline methods.

\noindent \textbf{Evaluation Metrics.} We consider three metrics to evaluate the proposed FedPC system in terms of clients' objectives, generalizability, and compatibility with new clients. 
\begin{enumerate*}[label=(\roman*)]
\item Performance of the current client model $\omega_c$ on the local dataset $X_c$. 
\item Performance of the current client model $\omega_c$ on other clients' datasets $\{\dots,X_{c-1}, X_{c+1} ,\dots\}$, i.e., the performance of the model $\omega_c$ without training on other clients and testing directly. For FedAvg and FedProx, it is the accuracy of the global model on all client test datasets.
\item Performance of the current client model $\omega_c$ on new clients, and the performance after single or multiple gradient descents, which can be considered as a meta-learning approach~\cite{fallah2020personalized}.
\end{enumerate*}

\begin{figure}[t]
\centering
\subfloat[StateFarm]{\includegraphics[width=0.5\linewidth]{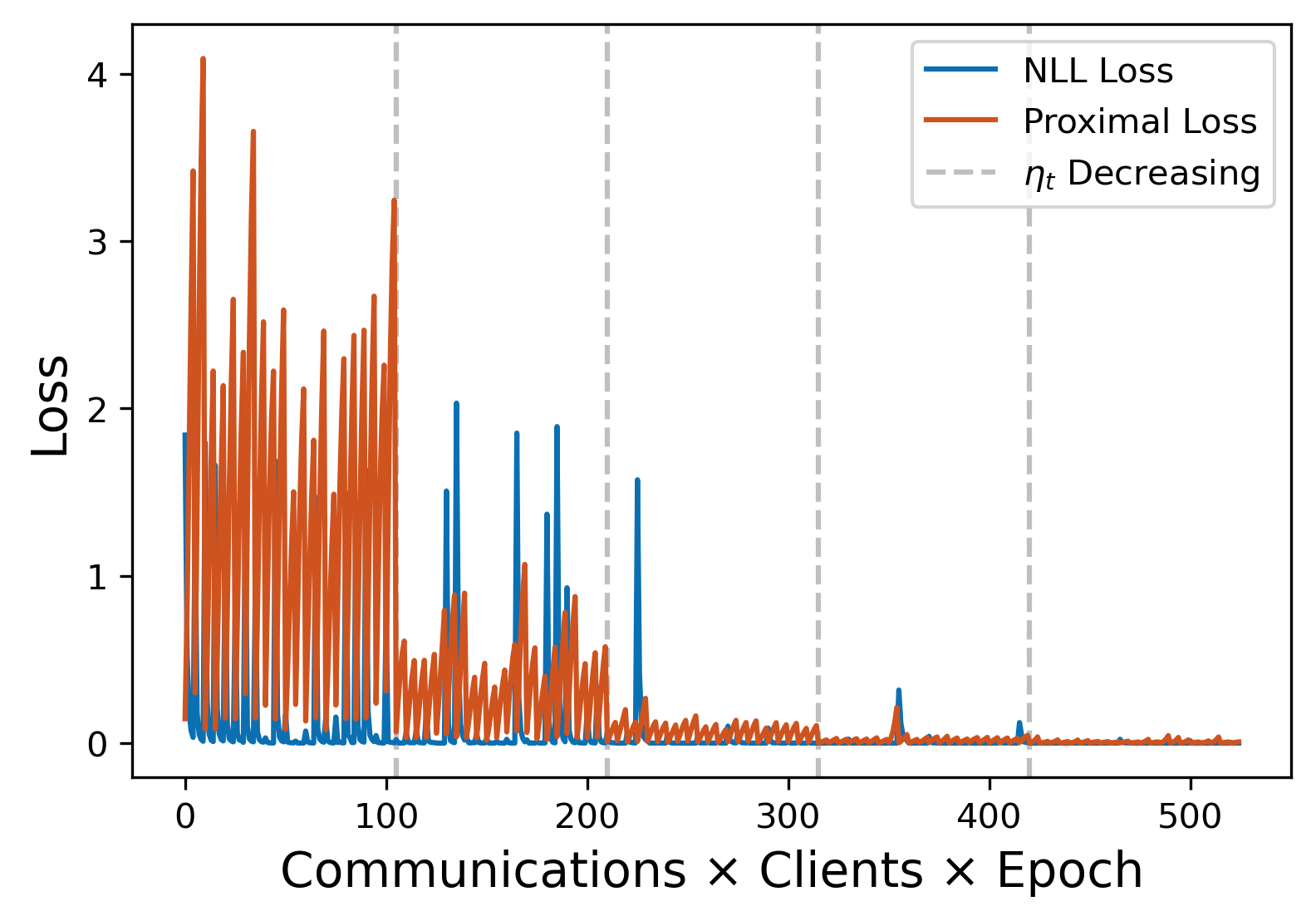}%
\label{Fig. StateFarm Loss}}
\hfill
\subfloat[AICity]{\includegraphics[width=0.5\linewidth]{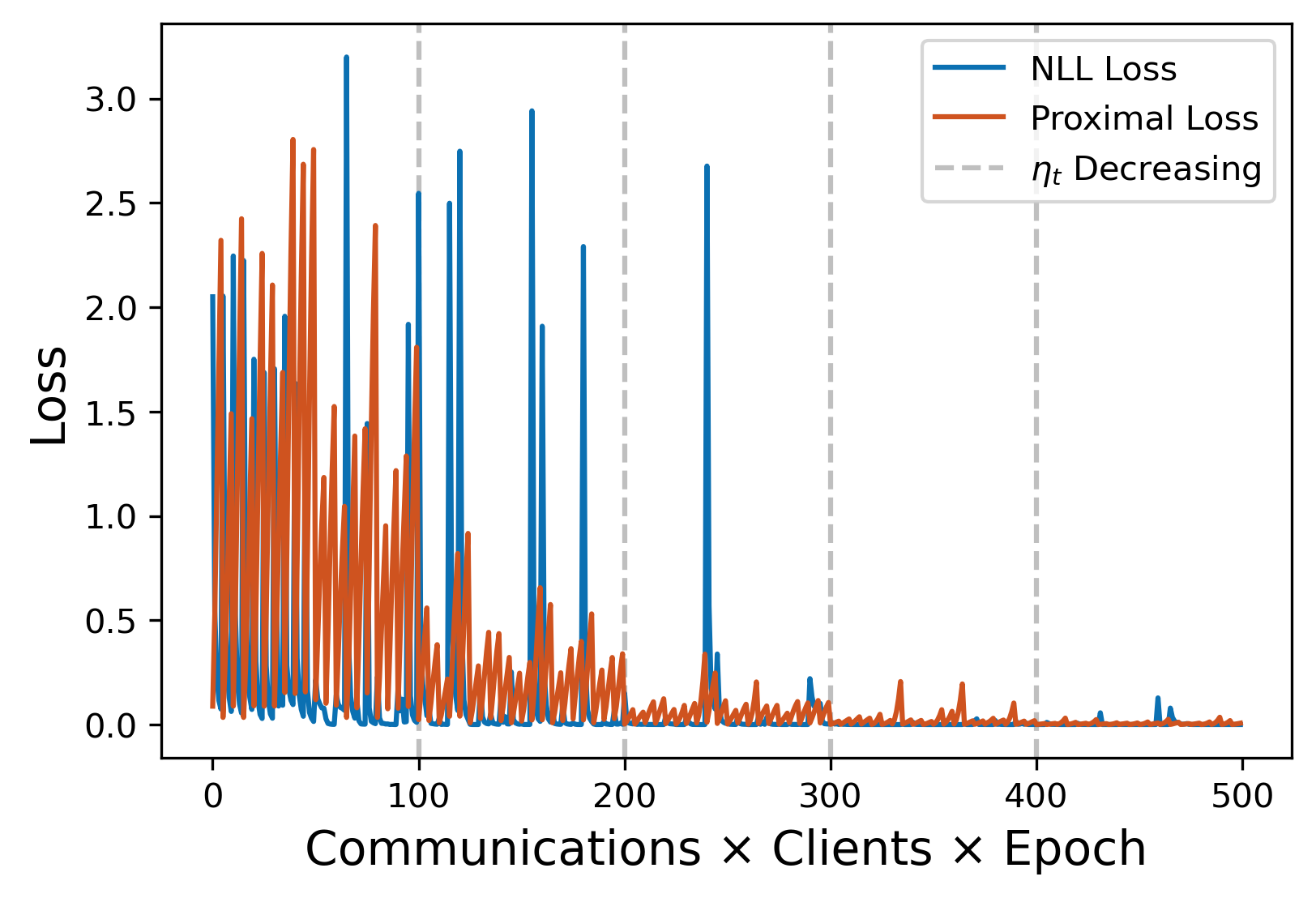}%
\label{Fig. AICity Loss}}
\caption{Convergence of FedPC on two datasets, (a) StateFarm and (b) AICity. Two loss functions, NLL loss $\mathcal{L}_{nll}$ and proximal loss $\mathcal{L}_{p}$, are expressed in \cref{Eq. loss function}. $\eta_t$ is the decreasing learning rate with the number of iterations.}
\label{Fig. Loss} 
\end{figure}

\begin{figure*}[t]
\small
\centering
\begin{tabular}{cccc}
~ & \textbf{\hspace{15pt} Metric (i) Client Objective} & \textbf{\hspace{15pt} Metric (ii) Generalizability} & \textbf{\hspace{15pt} Metric (iii) New Client} \\
\rotatebox[origin=c]{90}{\hspace{7pt} \textbf{StateFarm}} &
\raisebox{-0.5\height}{\includegraphics[width=0.3\linewidth]{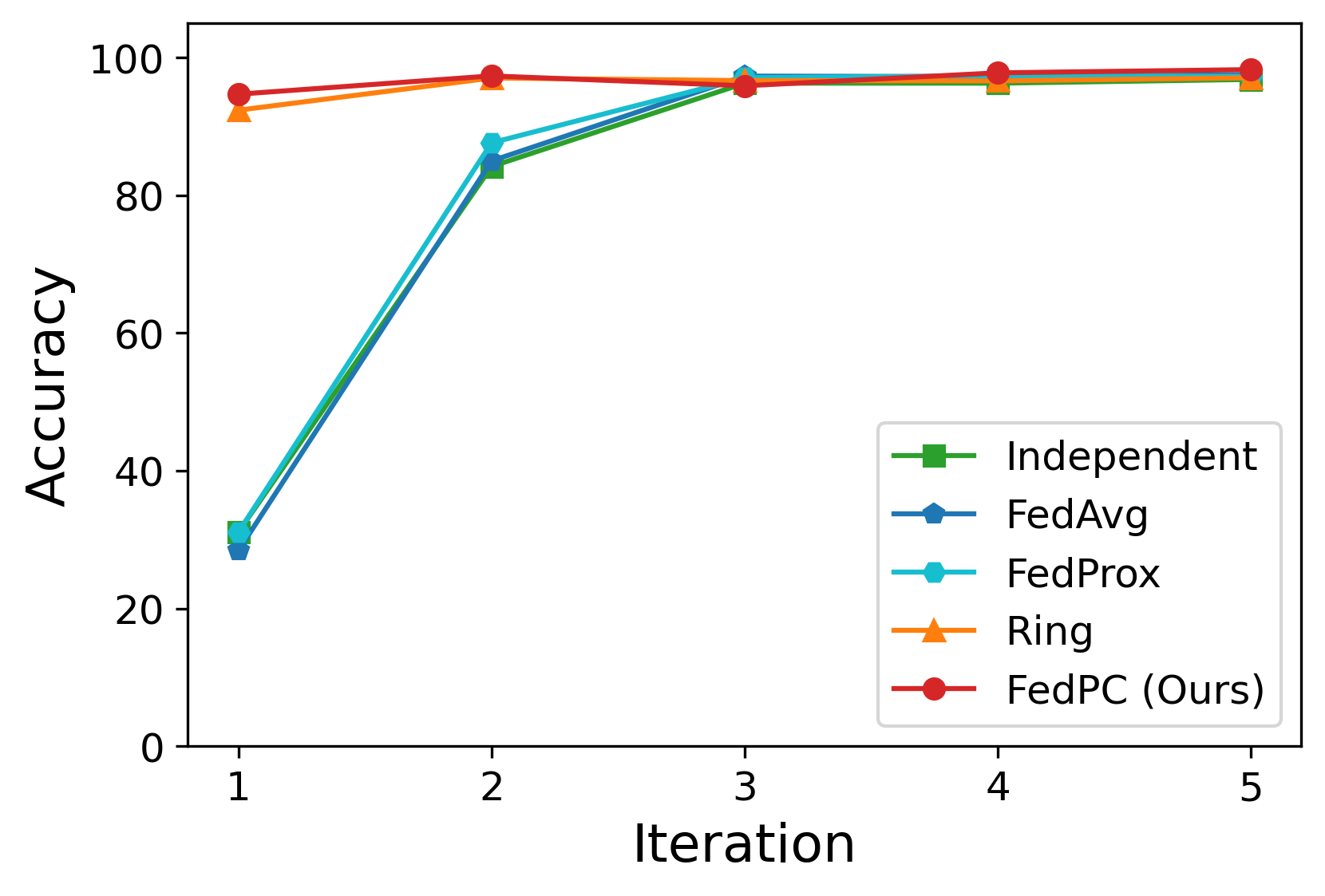}%
\label{Fig. SF_i}} &
\raisebox{-0.5\height}{\includegraphics[width=0.3\linewidth]{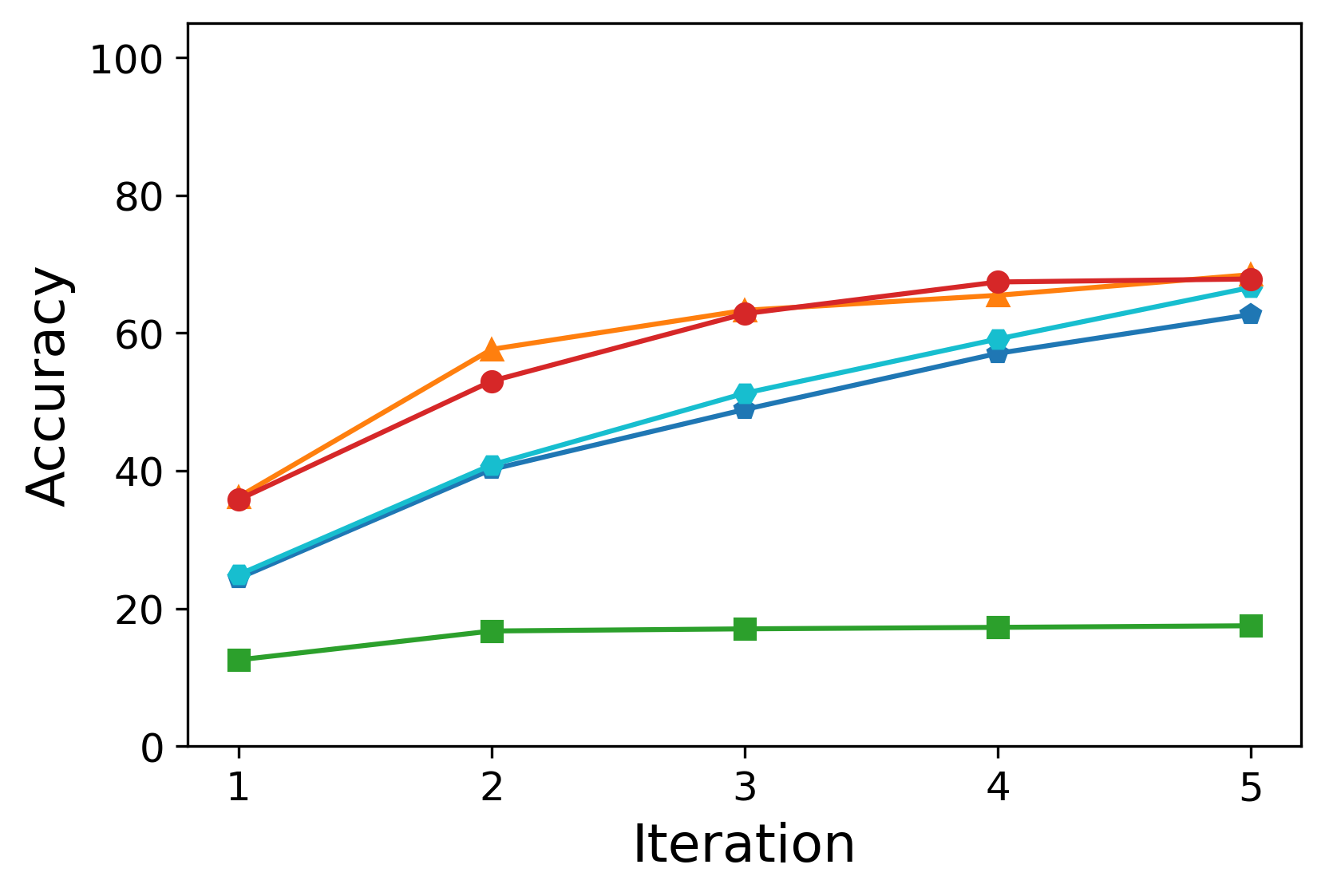}%
\label{Fig. SF_ii}} &
\raisebox{-0.5\height}{\includegraphics[width=0.3\linewidth]{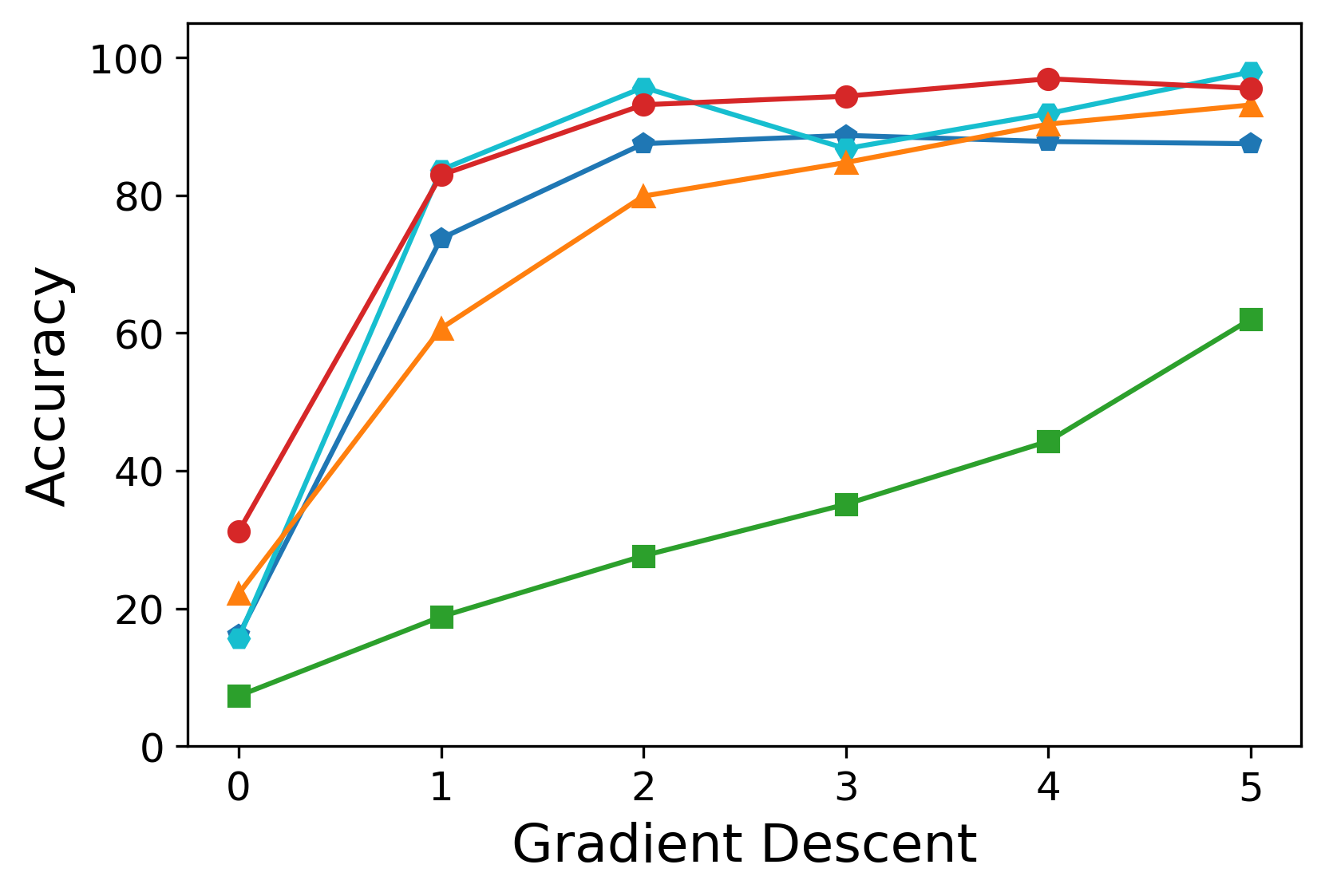}%
\label{Fig. SF_iii}} \\
\rotatebox[origin=c]{90}{\hspace{7pt} \textbf{AICity}} &
\raisebox{-0.5\height}{\includegraphics[width=0.3\linewidth]{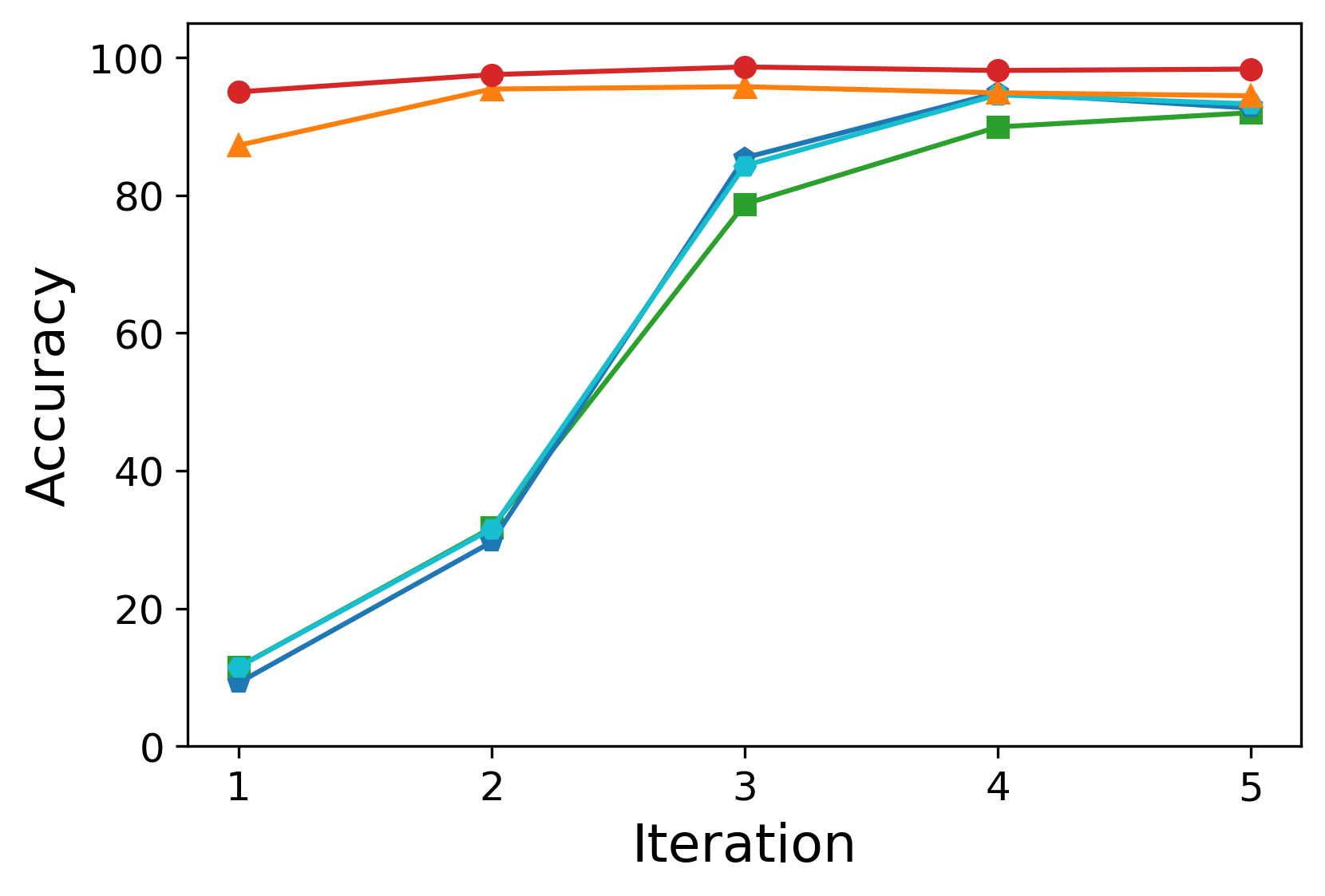}%
\label{Fig. AICity_i}} &
\raisebox{-0.5\height}{\includegraphics[width=0.3\linewidth]{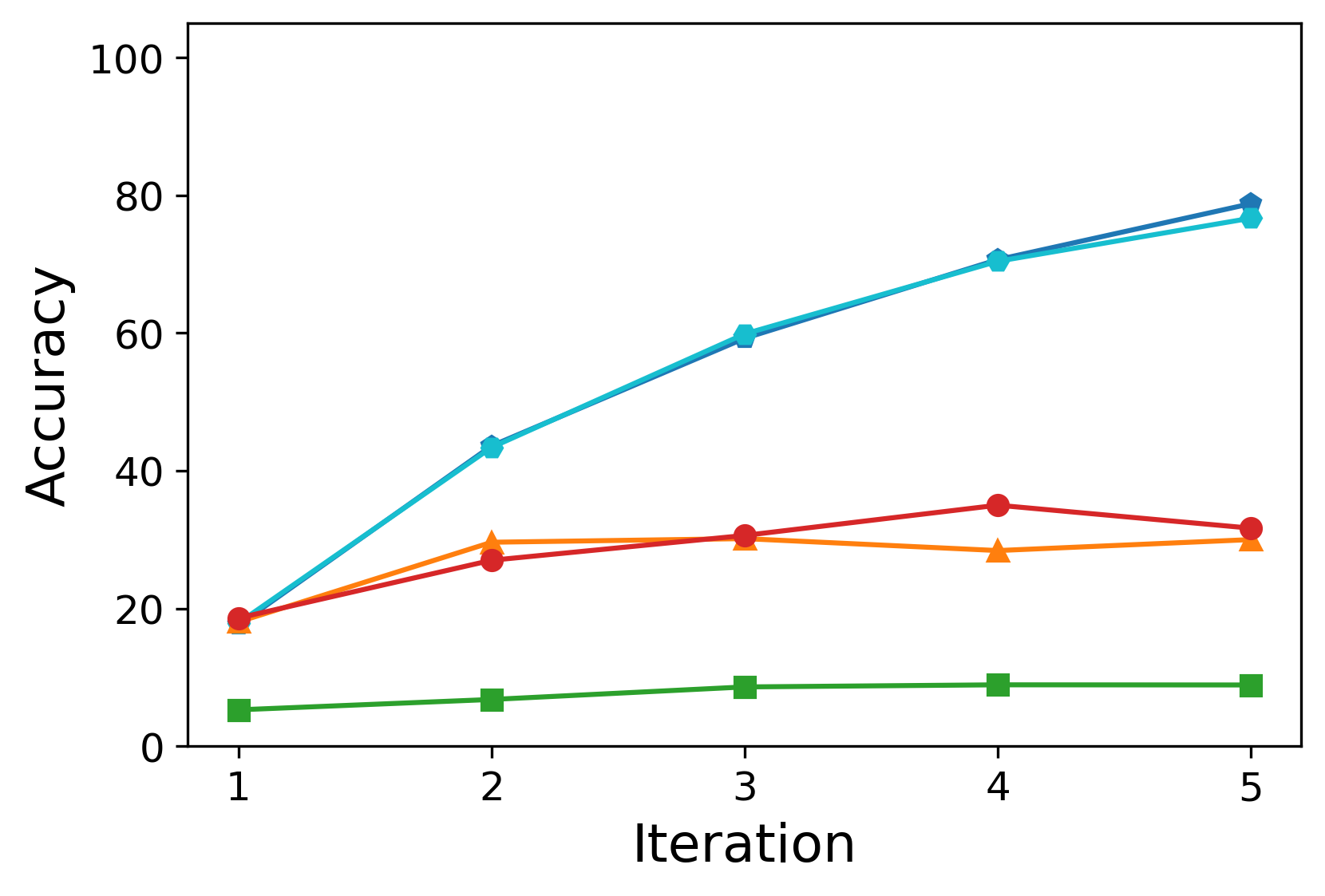}%
\label{Fig. AICity_ii}} &
\raisebox{-0.5\height}{\includegraphics[width=0.3\linewidth]{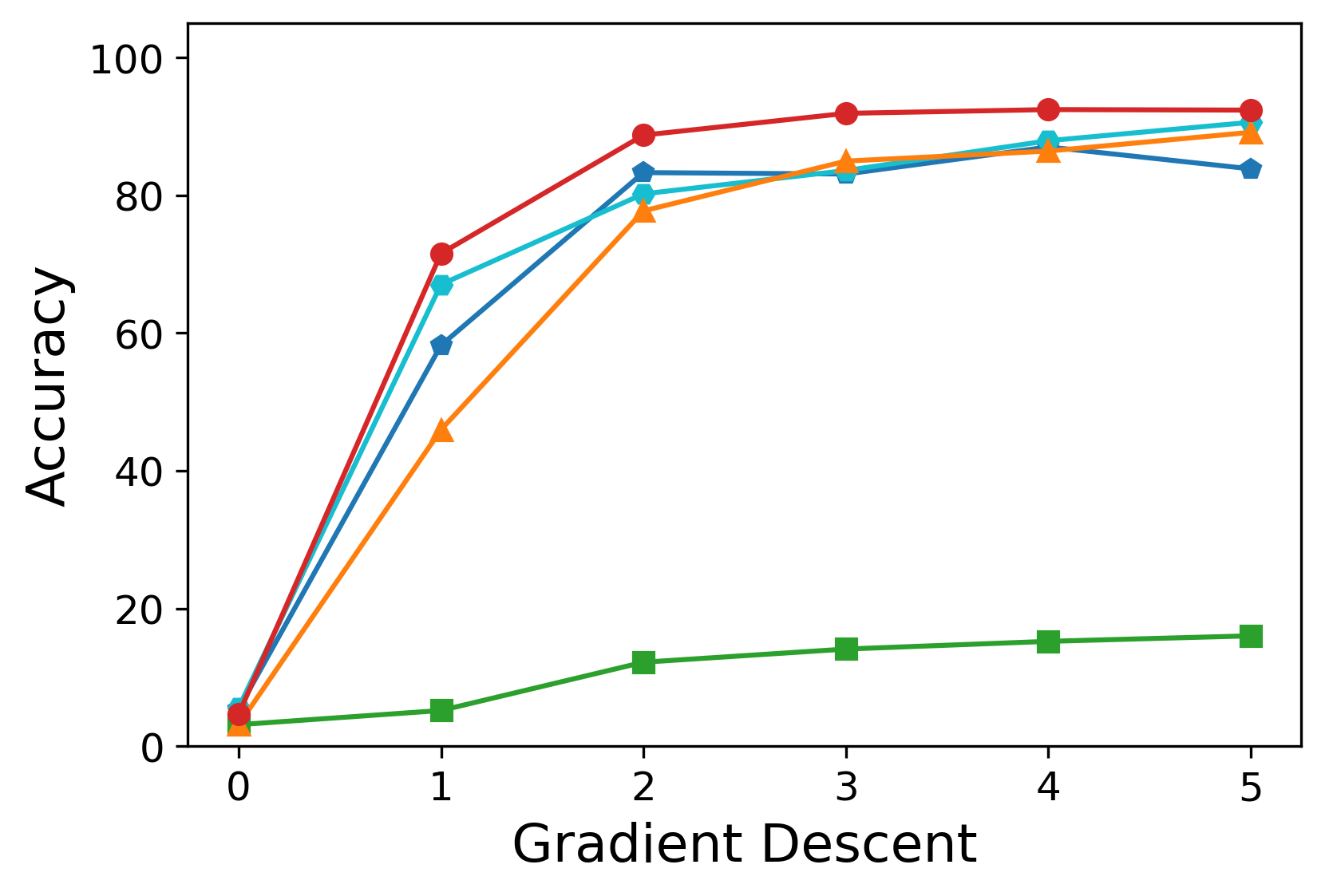}%
\label{Fig. AICity_iii}}
\end{tabular}
\caption{Experimental results on two data sets and three evaluation metrics. The evaluations are performed on the unseen test data set. The data points represent the average of the clients' results.}
\label{Fig. Experimental results} 
\end{figure*}

\subsection{Results}
\cref{Fig. Loss} depicts the convergence of the proposed FedPC system during training on both datasets. As explained in \cref{Sec:Training Strategy}, the optimization process involves a trade-off between two losses, namely NLL and proximal. Initially, the model $\omega_c$ is exposed to a high learning rate and NLL loss, causing it to diverge significantly from the previous client's model $\omega_{c-1}$. As training progresses and the learning rate decreases, the model gradually fine-tunes its parameters to strike a balance between accuracy and generalization. FedPC exhibits similar loss curves and convergence behavior on both datasets.

Subsequently, the proposed FedPC is thoroughly compared with four baselines on three evaluation criteria, as shown in \cref{Fig. Experimental results}.

\noindent \textbf{Metric (i) Client Objective.} Metric (i) evaluates the accuracy of the current client model on the current local dataset, which can be considered as assessing the convergence during the training of an isolated client. The discrepancy between independent learning, C2S FLs, and P2P FLs can be attributed to the differences in their initial models, leading to significant variances in the first few iterations. For both independent learning and C2S FLs, the initial models of all clients are pre-trained on ImageNet in the first iteration. On the contrary, for P2P FLs, only the initial model of the first client is pre-trained on ImageNet in the first iteration, while the initial models of all other clients are from their previous clients. As a result, the P2P FL frameworks initiate the knowledge dissemination process early on to expedite model convergence, without relying on aggregation as in C2S FLs.

\noindent \textbf{Metric (ii) Generalizability.} Metric (ii) evaluates the accuracy of the current client model on other client datasets, i.e., the assessment of the generalization ability of the model. For P2P FLs, this metric represents the average accuracy of a randomly selected client model on other clients, while for C2S FLs, it corresponds to the accuracy of the global model. It can be observed that P2P FLs exhibit comparable and notable generalizability using a randomly selected client model, even without focusing a global objective or performing aggregation. The differing results observed in the two datasets can be attributed to their varying degrees of data heterogeneity. The heterogeneity of the StateFarm dataset arises from the combination of multiple large discrete distributions, which include different vehicle models, as shown in Fig. 3. Consequently, a single aggregated global model cannot effectively handle the non-IID dataset. In contrast, P2P FLs yield higher average accuracy due to the ability of client models to achieve better accuracy on similar clients. The system heterogeneity of the AICity dataset is smaller, and its statistical heterogeneity is larger, which is reflected by a high variance distribution. Therefore, P2P FLs are only compatible with nearby models, leading to lower average accuracy.

\noindent \textbf{Metric (iii) New Client.} Metric (iii) evaluates the compatibility and generalization of the system for new clients. It can be observed that all frameworks exhibit limited compatibility with new clients, however, one or several gradient-descent personalization processes can yield comparable performance for new clients. The convergence rate of the models in the personalization process also reflects the distance between each system's output model and the optimal model for the new clients. It can be seen that the proposed FedPC has a slight advantages in terms of convergence speed.

\noindent \textbf{FedPC vs C2S FL.} From the three evaluations above, FedPC and C2S FL have their own advantages. On the one hand, the FedPC system can execute the knowledge dissemination and convergence process more rapidly without waiting for the server to perform the aggregation operation. Simultaneously, FedPC reduces server communication, computational, and storage overheads and eliminates privacy, security, and fairness issues on the server side. On the other hand, FedPC and C2S FL exhibit different generalization capabilities over two datasets. The generalization properties of FedPC are not expected to surpass those of C2S FL with aggregation operations, due to the fundamental differences in the objectives of the two systems. For FedPC, the locally trained model not only possesses the knowledge of other clients but also can be directly deployed locally. In contrast, for C2S FL, the local model lacks knowledge from other clients, and the global model does not perform well on the local dataset.

\section{Discussion}
\label{Sec:Discussion}
Although the experimental datasets used in this study were collected in the real world, certain information is missing, and the proposed FedPC can only be simulated through alternative methods. This section aims to discuss potential deployments to FedPC in real-world application scenarios.

\noindent \textbf{Pre-clustering.} Pre-clustering clients in FedPC networks using a priori information is one of the potential solutions for personalization, accelerating iteration, and improving accuracy. The proposed FedPC may encounter endless delays when the number of clients is infinite, and the iteration rate of the whole system will be limited to the training rate of the clients, which is the same concern in C2S FL frameworks. Hence, some works have proposed clustered FL to reduce iteration complexity and initialize personalization through clustering of similar models~\cite{sattler2020clustered}. However, for FedPC, data-driven or model parameter-driven clustering is not available since information about all clients on the same device is not available at a particular time. Thus, scenario information-driven FedPC clustering is a viable potential solution in real-world application scenarios. Clustering can be based on information related to vehicle type, vehicle make, driving scenario, driving habits, etc. Furthermore, clustering can eliminate system heterogeneity arising from different camera angles, resolutions, sampling rates, and vehicle interiors. Therefore, clustering can be considered as a pre-personalization process to further enhance the iteration rate, latency, accuracy, and robustness of the FedPC framework.

\noindent \textbf{Communication Protocol.} The FedPC framework with different communication protocols is designed to reduce the communication overhead for different application scenarios. VANET-based model propagation has two properties, highly dynamic and transmission range-limited. For highly dynamic vehicle application scenarios, VANET-based model transmission is also highly dynamic. Neighboring vehicle model transmission can significantly reduce communication overhead. Although the proposed FedPC employs the gossip protocol to simulate the randomness of vehicle connections, there are artificially random connections for real-world scenarios. First, vehicles in the same driving scenario are connected more frequently, e.g., vehicles in the same city, which can also be considered as a driving scenario clustering. Second, active vehicles are connected more frequently, e.g., taxis that are driven for extended periods daily. Third, the same vehicle manufacturers can have higher connection frequency, which eliminates system heterogeneity due to the consistency of in-vehicle devices. Therefore, the FedPC framework in real driving scenarios needs to consider vehicle connectivity heterogeneity to adopt appropriate communication protocols.

\noindent \textbf{Parallel Propagation of Multiple Models.} In this paper, we investigate the utilization of continual learning as a means of propagating knowledge. However, in real-world application scenarios, it may be possible to interconnect multiple vehicles and simultaneously propagate multiple models. Therefore, the aggregation paradigm is also a viable option for a multi-peer-to-peer protocol. In real-world application scenarios, a hybrid approach that combines both continual and aggregation as a knowledge dissemination scheme has the potential to more effectively leverage multiple models.

\section{Conclusion}
\label{Sec:Conclusion}

In this paper, we propose FedPC, a novel P2P FL approach for NDAR tasks that combines continual learning with a gossip protocol to propagate knowledge among clients. FedPC focuses on and satisfies each client's objective, enabling every client to have a personalized model. We evaluate the performance of FedPC on two real-world NDAR datasets, demonstrating its rapid knowledge dissemination, comparable generalizability, and swift compatibility with new clients. Furthermore, its low overhead in terms of communication, computation, and storage makes it well-suited for deployment in real vehicles.

Future work aims to further advance the deployment of FedPC in real-world application scenarios, including pre-clustering, VANET-based communication, incentive strategies, etc., to address potential issues such as data heterogeneity, high-dynamic connectivity, communication overhead, packet loss, latency, free-riding attacks, etc. One possible scenario involves the connection and propagation of models between vehicles driving in the same direction on a highway. Differences in vehicle speeds will lead to varying connection frequencies, making vehicle connections more realistic in real-world scenarios compared to the gossip protocol. A significant challenge in real-world applications is the reluctance of users to participate in knowledge dissemination due to concerns regarding safety, privacy, and communication costs. The lack of motivation among clients emphasizes the urgent need to address incentive strategies.

\newpage

{\small
\bibliographystyle{ieee_fullname}
\bibliography{egbib}
}

\end{document}